# Empirical Analysis of AI-based Energy Management in Electric Vehicles: A Case Study on Reinforcement Learning

Jincheng Hu, Yang Lin, Jihao Li, Zhuoran Hou, Dezong Zhao, *Senior Member, IEEE*, Quan Zhou, *Member, IEEE*, Jingjing Jiang, *Member, IEEE*, and Yuanjian Zhang, *Member, IEEE*

*Abstract*—**Reinforcement learning-based (RL-based) energy management strategy (EMS) is considered a promising solution for the energy management of electric vehicles with multiple power sources. It has been shown to outperform conventional methods in energy management problems regarding energy-saving and real-time performance. However, previous studies have not systematically examined the essential elements of RL-based EMS. This paper presents an empirical analysis of RL-based EMS in a Plug-in Hybrid Electric Vehicle (PHEV) and Fuel Cell Electric Vehicle (FCEV). The empirical analysis is developed in four aspects: algorithm, perception and decision granularity, hyperparameters, and reward function. The results show that the Off-policy algorithm effectively develops a more fuel-efficient solution within the complete driving cycle compared with other algorithms. Improving the perception and decision granularity does not produce a more desirable energy-saving solution but better balances battery power and fuel consumption. The equivalent energy optimization objective based on the instantaneous state of charge (SOC) variation is parameter sensitive and can help RL-EMSs to achieve more efficient energy-cost strategies.**

## I. INTRODUCTION

THE automotive industry is gradually phasing out conventional vehicles using fossil fuels as a single energy source in the context of global fossil energy depletion and global warming. Multi-power source electric vehicles (MPS-EVs) are considered to be an eco-friendly substitute to achieve green-mobility in the future,

*Corresponding author: Yuanjian Zhang*

Jincheng Hu, Jihao Li, Jingjing Jiang and Yuanjian Zhang is with the Department of Aeronautical and Automotive Engineering, Loughborough University, Leicestershire, LE11 3TU, United Kingdom (e-mail: J.Hu2@lboro.ac.uk; J.Li4-21@student.lboro.ac.uk; J.Jiang2@lboro.ac.uk; Y.Y.Zhang@lboro.ac.uk).

Yang Lin and Zhuoran Hou are with College of Automotive Engineering, Jilin University, Changchun 130022, China (e-mail: liny21@mails.jlu.edu.cn; houzr20@mails.jlu.edu.cn).

Dezong Zhao is with the James Watt School of Engineering, University of Glasgow, Glasgow, United Kingdom (e-mail: Dezong.Zhao@glasgow.ac.uk).

Quan Zhou is with the Department of Mechanical Engineering, University of Birmingham, Birmingham B15 2TT, United Kingdom (e-mail: q.zhou@bham.ac.uk).

Color versions of one or more of the figures in this article are available online at http://ieeexplore.ieee.org

which introduce sustainable powertrain solutions based on clean electricity and hydrogen energy to improve energy efficiency. However, it cannot maximize the sustainability of future travel through innovations in the powertrain configuration alone. Energy management strategy (EMS) plays another important role in comprehensively releasing the potential of energy-saving [1].

Compared to the technical difficulty and high cost of hardware development, EMS for MPS-EV as an on-board control software has the advantage of low development cost and short development cycle. The research of EMS has made significant progress in the past decades. The EMS that has been proposed and applied can be classified into rule-based approaches, global optimization-based approaches, instantaneous optimization-based approaches, and artificial intelligence-based (AI-based) approaches. Rule-based EMS can be divided into deterministic rule-based strategies [2, 3], fuzzy rule-based strategies [4, 5], and filter-based strategies [6]. Rule-based EMS is easily understood, highly interpretable, and has good real-time performance. However, it heavily relies on expert knowledge and frequently exhibits worse energy efficiency [7]. Global optimization-based methods, represented by Dynamic Programming (DP) [8], Genetic Algorithms (GA) [9], and Particle Swarm Optimization (PSO) [10], can plan an operating trajectory with optimal performance based on the known driving cycle. They can achieve the best results in theoretical calculations but are challenging to apply in the real-time applications. Instantaneous optimization-based methods are constructed to minimize energy consumption under single or multi-step sequences, such as Equivalent Consumption Minimization Strategy (ECMS) [11] and Model Predictive Control (MPC) [12]. They reduce the real-time computational cost and achieve energy-saving performance close to global optimization approaches but are highly dependent on environmental information. AI-based approaches are promising energy management solutions for MPS-EVs, which include machine learning and deep learning-based knowledge migration methods [13, 14, 15] and RL-based active control methods [16, 17]. The RL demonstrates impressive capabilities for robust regression analysis and strategy development, and the RL-based EMS has shown substantial potential for real-time optimal control in the energy management problem of MPS-EV [18].

RL is a class of AI methods for the actionism doctrine that explore and optimize behavioural policies based on the trajectory of the agent-environment interactions to maximize expected returns [19]. Theoretically, RL can provide accurate state-value estimations based only on a large number of trial-and-error processes without knowledge of the environment. As a highly anticipated approach, RL-based EMS has been widely applied to energy management problems in MPS-EVs. Xu et al. used Q-learning to train RL-EMS in a parallel HEV model. The result was an 8.89% improvement in fuel economy compared to a logic-based control strategy and a 0.88% improvement in fuel economy compared to an ECMS [20]. Lin et al. used Q-learning-lambda to train RL-EMS in HEV and PHEV models and compared it with rule-based EMS under ten different operating conditions, and the results showed that the average fuel consumption decreased by 28.8% in HEV and 30.4% in PHEV [21]. Chen et al. used the SARSA algorithm to solve the energy management problem. Experimental results under real-world driving conditions showed that the proposed method could significantly improve fuel economy by up to 21% compared to rule-based EMS [7]. With the rise of deep learning research, deep reinforcement learning has also exhibited reasonable control optimization capabilities. Deep reinforcement learning (DRL), represented by DQN, has also been applied to RL-EMS, which is superior in real-time and fuel economy compared to the results of rule-based EMS and DP [22, 23]. RL algorithms have become a popular choice for current MPS-EV EMS development. However, the different algorithms on the effectiveness of MPS-EV energy management have not been adequately discussed in past studies, and the underlying reasons for the performance gap of algorithms in MPS-EV energy management problems have not been analyzed in depth.

The perception and decision-making capability of RL-based EMS are limited by the state and action space in the vehicle simulation. The state and action space are continuous in the energy management problem of MPS-EV. Discretizing the continuous space is a standard solution for RL-EMS. Hu et al. used $(T_{dem}, SOC)$ as the state and $T_e$ as the action; $T_{dem}$ is the vehicle demand torque; $SOC$ is the current battery charging state; $T_e$ is the engine output torque. They discretized the continuous space into 24 values [22]. Liu et al. used $SOC$ as the state with a discrete interval of 0.013; motor torque $T_m$ and engine torque $T_e$ as the action, and the action space is discretized into 20 values [24]. Yuan et al. discretized the state to 5 values and action discretized to 200 values [25]. Other common state-action discretization granularities are 8 [26], 24 [27], and 50 [28]. The discrete granularity of the state and action space in the RL-EMS algorithm can significantly affect the state migration process in optimal control. No reasonable discrete granularity criteria have been provided in previous studies.

Common optimization objectives of EMS for MPS-EV energy management optimal control include fuel consumption [29, 30], fuel economy [31], battery life [32, 33], emissions [34, 35], and the reward function in reinforcement learning-based MPS-EV energy management schemes are taken directly from the optimization objectives of energy management optimal control. For example, the reward function is directly composed of the dollar-settled fuel and battery energy usage costs [24]; Xiong et al. proposed the total energy loss of a PHEV as the reward function [36]; Liu et al. used fuel consumption as the reward function for reinforcement learning [37]; Kouche-Biyouki et al. in the energy management problem of an FCEV used electricity consumption to construct the reward function [38]. In previous studies, the reward function is given directly, and the rationality of the internal design of the reward function is not fully explained. In addition, the influence of the choice of different reward functions on the strategy of RL-EMS has never been analyzed.

Many hyperparameters can severely affect the learning process of RL, and this effect is often nonlinear and difficult to estimate. Deficient hyperparameter selection can lead to catastrophic policy training results. In past RL-EMS studies, hyperparameters were given directly, and only a few parameters were systematically analyzed. For example, Qi et al. discussed the training results of RL-EMS with different learning rate settings [39]; Qi et al. searched for the optimal value of exploration rate in the interval [0.1, 0.3, 0.5, 0.7, 0.9] [40]. The common situation is that most RL-EMS studies do not describe the basis for setting the hyperparameters of the reinforcement learning algorithm and the exploration process and cannot provide experience for related studies.

In summary, most of the past research has been devoted to improving fuel-economy or energy-saving under specific driving conditions and vehicle settings by applying the RL methods. However, they do not accurately detail the basic principles of RL-EMS design and the RL-EMS design paradigm in MPS-EV energy management problems. It is not conducive to promoting advanced RL methods in the energy management of MPS-EV. We develop an empirical analysis of the critical factors affecting RL-EMS performance to address this gap. In this paper, two typical MPS-EV models (PHEV and FCEV) are constructed as research cases to analyze the effects of different algorithms, perception-decision granularity, reward functions and hyperparameter settings on RL-EMS in terms of powertrain efficiency, average fuel consumption and amount of battery charge state variation. Based on the experimental results, this paper describes the influence of the above four critical factors on the control effect of RL-EMS. The contributions of this paper are as follows:

In this paper, we analyze the impact of reinforcement learning algorithms on RL-EMS training results in terms of the action-state value (Q-value) update principles of temporal difference, Off-policy, and multi-step update. The potential factors for the difference in energy consumption optimization of different RL-EMS solutions are studied. The experiments implement four energy management strategies based on different reinforcement learning algorithms: MC, Q-learning, SARSA, SARSA-Lambada. The specific effects of the three Q-value updating methods on RL-EMS are demonstrated by presenting the SOC trajectory and the powertrain working diagram of RL-EMS based on different reinforcement learning algorithms during training.

This paper evaluates the effect of perception-decision granularity on the effectiveness of RL-EMS and explains the underlying reasons for this performance difference. In the experiments, RL-EMS is constructed based on the Q-learning algorithm, and the state-action space granularity is varied in the constructed RL-EMS to analyze the role of perception-decision granularity in applications by comparing the control effects under different granularity settings.

This paper analyzes the specific effects of the primary hyperparameters on the effectiveness of RL-EMS. The analysis results with generality can guide the hyperparameter tuning of RL-EMS. In the experiments, the hyperparameter selection space of RL-EMS is listed, and the experimental results reveal the influence of hyperparameters on RL-EMS in MPS-EV energy management problems by analyzing the RL-EMS training effects under different hyperparameter configurations.

This paper investigates the influence of reward function design and its parameters on the optimization performance of energy saving. In the experiments, RL-EMS calculates the total energy consumption of MPS-EV from two perspectives of instantaneous consumption and current-overall consumption. It reveals the role of reward function design and its parameters in RL-EMS applications by comparing the experimental results of different reward functions.

The remaining sections of the paper are organized as follows. Section II describes the two configurations and mathematical models used in this paper and explains the general process of multi-power source vehicle energy management optimization. Section III introduces the basic concepts of RL-EMS and describes the experimental setup of this paper. Section IV shows the results of the experiments and gives the analysis. Section V draws conclusions from the experiment.

## II. ENERGY MANAGEMENT PROBLEM IN MPS-EVs

### 2.1 Electric Vehicles with Multiple Power Sources

A typical electric vehicle with multiple power sources can supply or store electricity during the driving process through multiple energy storage devices. Fig. 1 shows the common configuration of MPS-EVs. By removing some control nodes and energy storage devices, a variety of MPS-EV configurations can be derived: hybrid electric vehicle (HEV), plug-in hybrid vehicle (PHEV), and fuel cell vehicle (FCEV). Fossil-fuel combined MPS-EVs such as hybrid electric vehicles (HEV) and plug-in hybrid electric vehicles (PHEV) can enhance fuel utilization efficiency by extracting power from multiple energy storage devices, unlike internal combustion engine (ICE) vehicles. In the case where only the motor is adopted as the drive component, zero-emission MPS-EVs, represented by fuel cell vehicles (FCEV), can fully use clean energy to extend the range of electric vehicles and optimize the efficiency of batteries, unlike battery electric vehicles (BEV).

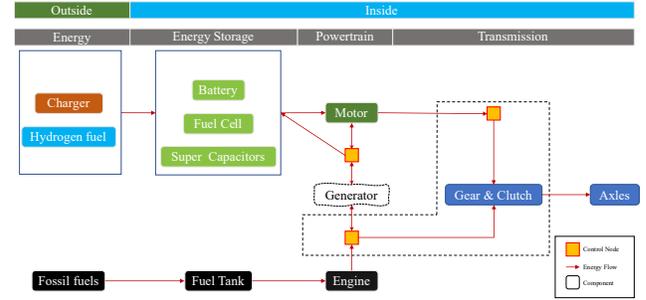

**Fig .1.** Multi-power sources configuration in Electric Vehicle.

### 2.2 Mathematics Models for Energy Management in EVs

Although real-world vehicle tests can provide reliable performance analysis data, it is unacceptably costly in terms of both economy and time. Therefore, the development of MPS-EV control systems often relies on numerical modelling and simulation method. Vehicle power components and transmission structures are the most crucial represented objects in numerical modelling, which is intended to reflect the characteristics and basic operating principles of the components. In this paper, a kinematic approach, also known as the backward approach, is used to build PHEV and FCEV simulation models for calculating the physical quantities required for the vehicle simulation process. The PHEV and FCEV configurations used are shown in Fig. 2 and Fig. 3, and the relevant parameters are shown in Table 1 and Table 2.

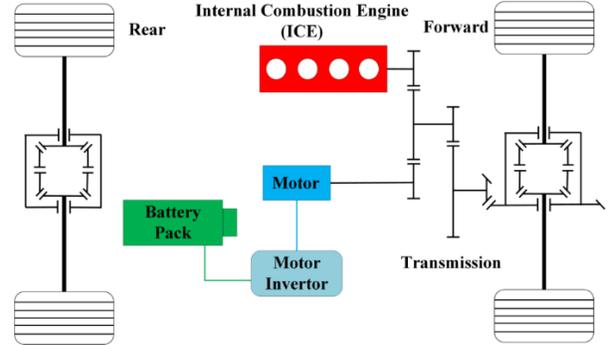

**Fig. 2.** The schematic of the PHEV configuration.

TABLE I
PHEV Model Parameters.

| Parameters | Value |
| --- | --- |
| Curb weight | 1200 kg |
| Engine max torque | 165 Nm |
| Engine max power | 102 kW |
| Engine max speed | 6500 rpm |
| EM max torque | 307 Nm |
| EM max power | 126 kW |
| Transmission gear number | 5 |
| Transmission gear ratio | 1st:3.527, 2nd:2.025, 3rd:1.382, 4th:1.058，5th:0.958 |
| Final differential ratio | 4.021 |

| Battery capacity | 20.8 Ah |
| Wheel radius | 2.2 m |

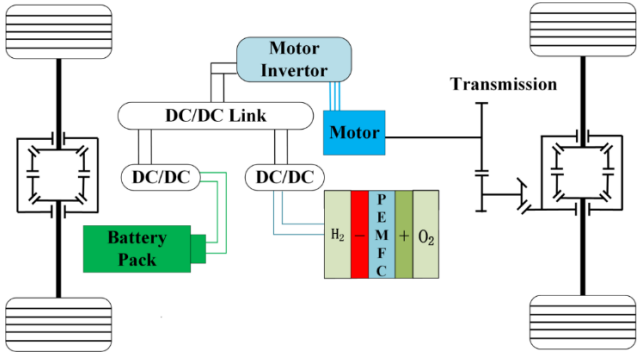

**Fig. 3.** The schematic of the FCEV configuration.

TABLE II
FCEV Model Parameters.

| Parameters | Value |
| --- | --- |
| Curb weight | 1200 kg |
| Motor max torque | 2500 Nm |
| Motor max speed | 3500 rpm |
| Motor max power | 249 kW |
| Battery voltage | 600 |
| Battery capacity | 88 Ah |
| Battery internal resistance | 0.06317 |
| Fuel cell maximum power | 55 kW |
| Wheel rolling radius | 0.32m |
| Differential ratio | 7.38 |

### 2.2.1 Vehicle Dynamic Model

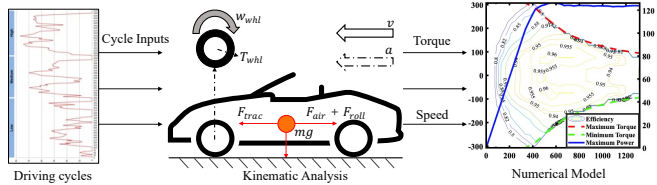

**Fig. 4.** The schematic of kinematic approach.

In the backward model, the calculation process for the performance of vehicle longitudinal driving is shown in Figure 4. The current speed determines the forces acting on the wheels. Then the power and energy consumptions of the vehicle during longitudinal driving are calculated based on the transmission and the component efficiency parameters. The driving force acting on the wheels $F_w$ can be described as follows:

$$F_w = F_{trac} + F_{air} + F_{roll} + F_{gravity} \qquad (1)$$

$$\begin{cases} F_w = ma \\ F_{air} = \dfrac{1}{2}\rho C_d A v^2 \\ F_{roll} = \cos(\beta) f_{roll} mg \\ F_{gravity} = \sin(\beta) mg \end{cases} \qquad (2)$$

where $F_w$ is the tangential force, $m$ is the vehicle mass, $a$ is the current vehicle acceleration; $F_{air}$ is the air resistance, $\rho$ is the air density, $C_d$ is the air resistance coefficient, $A$ is the vehicle windward area, $v$ is the current vehicle speed; $F_{roll}$ is the vehicle rolling resistance, $g$ is the gravitational acceleration, $f_{roll}$ is the rolling resistance coefficient, $\beta$ is the road slope; $F_{gravity}$ is the gravity component.

It can be obtained according to forces acting on the wheels:

$$T_{whl} = \frac{F_{trac}}{r_{whl}} \qquad (3)$$

where $T_{whl}$ denotes the wheel torque and $r_{whl}$ denotes the wheel radius.

### 2.2.2 Engine in PHEV

In this paper, the dynamic and temperature rise characteristics of the engine of the PHEV are ignored, and the quasi-static characteristics of the engine are described by the Map, which describes the efficiency of the power component at different speeds and output torques. Fig. 5 shows the efficiency of the engine in the target PHEV model at different speeds and torques, providing visual curves depicting the motor maximum torque limiting function $f_{M_{limit}}$ and the engine maximum torque limiting function $f_{T_{ICE\_max}}$.

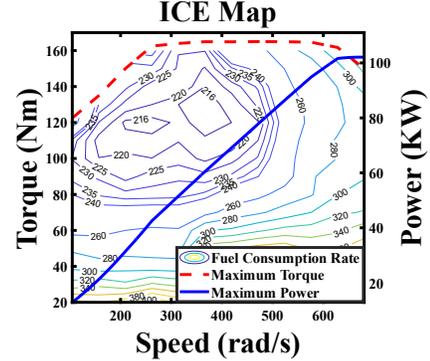

**Fig. 5.** The powertrain model in the PHEV. (a) Motor map; (b)ICE map.

The external characteristic curve constrains the torque of the engine during the simulation. Under a given speed, the engine torque cannot exceed the maximum engine torque $T_{ICE\_max}$, which depends on the current speed of the engine:

$$T_{ICE\_max} = f_{T_{ICE\_max}}(W_{ICE}) \qquad (4)$$

### 2.2.3 Fuel Cell in FCEV

The type of fuel cell introduced in this paper is the proton exchange membrane fuel cell (PEMFC).

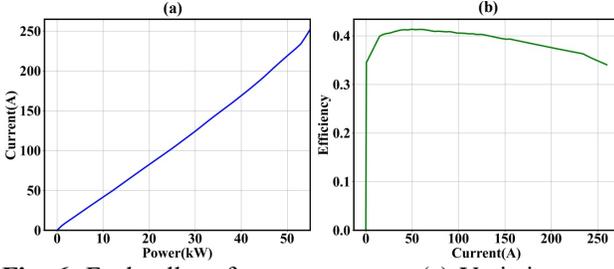

**Fig. 6.** Fuel cell performance curves. (a) Variation curves of the fuel cell power and current；(b) Variation curves of the fuel cell current and efficiency.

Due to its special physicochemical properties, the construction of an on-board fuel cell simulation model is very difficult. This paper constructs the model by fitted the fuel cell performance curves (Fig. 6) based on PEMFC experimental test data only. By retrieving the fuel cell characteristic curve, the current fuel cell operating current $I_{FC}$ and efficiency $e_{FC}$ can be quickly acquired based on the current fuel cell output power $P_{FC}$ :

$$I_{FC} = f_{PI}\left(P_{FC}\right) \tag{5}$$

$$e_{FC} = f_{eFC}\left(I_{FC}\right) \tag{6}$$

where $f_{PI}$ denotes the variation curve of the fuel cell power and current shown in Fig. 6(a); $f_{eFC}$ denotes the variation curve of the fuel cell current and efficiency shown in Fig. 6(b).

### 2.2.4 Motor Model

The dynamic and temperature rise characteristics of the motors are ignored in the experiment and the map is used to describe the quasi-static characteristics of the motors. The motor models for the FCEV and PHEV are described by the motor characteristics maps shown in Figure 7. $T_w$ , $w$ represent the torque and speed of the motor. When the vehicle is coasting, the wheels can drive the motor to rotate and the motor switches to generator mode, and the motor uses external energy to charge the battery. The motor also has corresponding maximum torque $T_{M\_max}$ and minimum torque $T_{M\_min}$ at the given speed, and this constraint must not be exceeded during the simulation:

$$T_{M\_max}, T_{M\_min} = f_{M\_limit}\left(W_M\right) \tag{7}$$

The efficiency of the motor can be obtained in Fig. 7:

$$\eta_M = f_{e_M}\left(T_M, W_M\right) \tag{8}$$

where $\eta_M$ denotes the motor efficiency; $T_M$ denotes the motor torque; $W_M$ is the motor speed; $f_{e_M}$ denotes the motor characteristics map in Fig. 7. Further, the motor power $P_M$ can be described as follows:

$$P_M = \begin{cases} \dfrac{T_M W_M}{\eta_M}, T_M \geq 0 \\ T_M W_M \eta_M, T_M < 0 \end{cases} \tag{9}$$

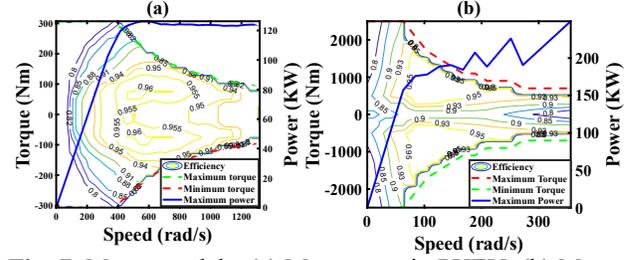

**Fig. 7.** Motor models: (a) Motor map in PHEV; (b) Motor map in FCEV.

### 2.2.5 Battery Model

In this paper, the battery type adopted by the PHEV and FCEV is LiFePO4 (LFP), and the battery model does not consider battery ageing and temperature variations. The temperature is fixed at 25 ° C.

The PHEV is equipped with a small battery pack with a total capacity $Q_{max}$ of 20.8 Ah. The open circuit voltage $V_{oc}$ and internal resistance $R_{int}$ of the battery can be acquired by interpolation based on the current battery state of charge. Fig. 8 shows the internal resistance curve and open circuit voltage curve of the PHEV battery model.

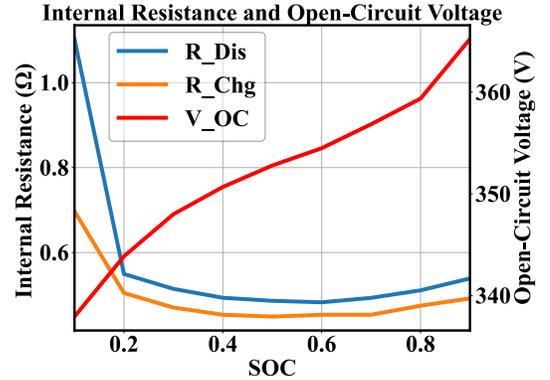

**Fig. 8.** PHEV Battery performance curves.

The FCEV uses a large-capacity battery with a capacity of up to 88Ah and a fixed internal resistance $R_{int}$ of $0.06317\,\Omega$ . The open circuit voltage $V_{oc}$ of this battery can be obtained by the open circuit voltage curve shown in Fig. 9 according to the SOC.

The battery is modelled through an equivalent circuit model, ignoring the effect of temperature on SOC. The current $I$ of the battery can be defined as:

$$I = \frac{V_{oc} - \sqrt{V_{oc}^2 - 4R_{int}P_{batt}}}{2R_{int}} \tag{10}$$

where $P_{batt}$ is output power of the battery. Based on the battery current $I$ and the battery capacity $Q_{max}$, the variation of state of charge $\Delta SOC$ can be calculated as:

$$\Delta SOC = -\frac{1}{Q_{max}} \tag{11}$$

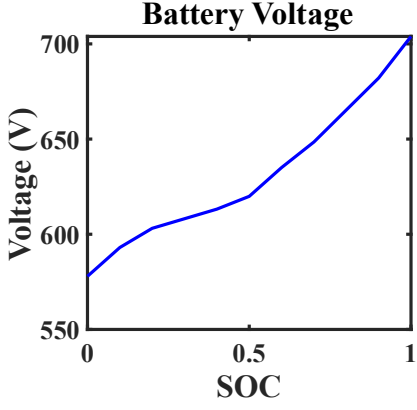

**Fig. 9.** FCEV battery performance curves.

### 2.3 General Energy Management Problem in MPS-EVs

The energy management problem for MPS-EV in the real-world is a multi-objective optimization problem. Considering constraints of multiple power sources, for which the generic optimization objective function can be described as follows:

$$J = \int_0^T L(t, u_t) dt \qquad (12)$$

where $u_t$ represents the current control signal; $t$ represents the current simulation time; $T$ represents the operating time of the EMS, which is the driving cycle duration. The energy consumption of the MPS-EV is set chiefly as the optimization objective. The energy consumption function $L$ can be expressed as:

$$L(t, u(t)) = \dot{m}_F(t, u_t) + \alpha \dot{m}_{E_q u_F}(t, u_t) + \beta L_{BAT}(\delta_e) \qquad (13)$$

where $\dot{m}_F$ represents the mass of fuel consumed at the moment; $\dot{m}_{E_q u_F}$ represents the equivalent fuel consumption of electricity consumed at the moment; $L_{BAT}$ represents the battery degradation cost and $\delta_e$ represents the current electricity consumption.

As a numerical optimization problem, energy management must consider various constraints in the real world. The most common constraints are that the battery state of charge cannot fall below a specified threshold and the battery cannot operate at a high current continuously, the torque provided by the engine and motor is bounded. In this paper, the following constraints need to be satisfied by all powertrain components in the experiments of the energy management problem:

$$\begin{cases} T_{ICE_{min}} \leq T_{ICE} \leq T_{ICE_{max}} \\ T_{M_{min}} \leq T_M \leq T_{M_{max}} \\ W_{ICE_{min}} \leq W_{ICE} \leq W_{ICE_{max}} \\ W_{M_{min}} \leq W_M \leq W_{M_{max}} \\ SOC_{min} \leq SOC \leq SOC_{max} \end{cases} \qquad (14)$$

where $T_{ICE}$ is the engine torque at any given moment; $T_M$ is the motor torque at any given moment; $W_{ICE}$ is the engine speed at any given moment; $W_M$ the motor speed at any given moment. The battery state of charge constraint and penalty function $\psi_\xi$ is expressed as:

$$\psi_\xi(SOC_t) = \begin{cases} w_{dis}(SOC_t - SOC_{max}), SOC_t > SOC_{max} \\ w_{chg}(SOC_t - SOC_{min}), SOC_t < SOC_{min} \\ 0, SOC_{min} \leq SOC \leq SOC_{max} \end{cases} \qquad (15)$$

where $SOC_t$ denotes the $SOC$ at moment $t$; $w_{dis}$ and $w_{chg}$ are the penalty functions for overcharging and over discharging, which are often set according to the needs of the research. Therefore, the objective function for MPS-EV energy management considering the battery charge state can be defined as:

$$J = \int_0^T L(t, u(t)) dt + \psi_\xi(\xi_T) \qquad (16)$$

### III. RL BASED EMS FOR MPS-EVs

#### 3.1 Reinforce Learning Method

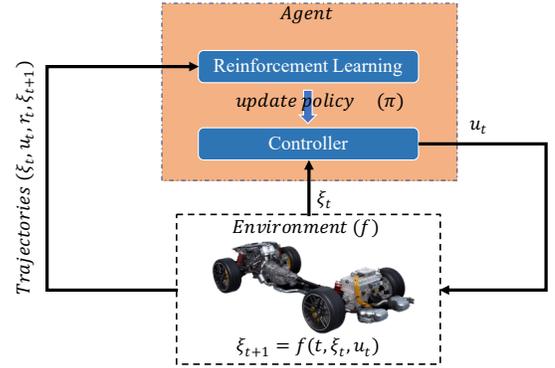

**Fig. 10.** The sequential decision making in MDP of RL-EMS

In the MPS-EV energy management problem, the on-board energy management system acts as the *Agent* and the RL-based EMS acts as the *policy* ($\pi$) to guide the agent to generate the control signal $u_t$ according to the current state $\xi_t$. The vehicle simulation based on the driving cycle is developed as the target environment, and Fig. 10 shows the control process to the vehicle model by the on-board energy management system. The simulation process satisfies the Markov Decision Process (MDP):

$$\xi_{t+1} = f(t, \xi_t, u_t) \qquad (17)$$

for any time $t$, $\xi_{t+1}$ is only related to $\xi_t$, independent of the state at other moments. The simulation cycle is one complete driving cycle. During the driving cycle, the physical quantities such as the speed and acceleration of the vehicle model change with time and the range of physical quantities is taken as the potential state space of the MDP [41]. In this paper, the physical quantities of the vehicle model are perfectly observable and the MDP process consists of the following five elements: $(X, A, P_{XA}, \gamma, R)$. $X$ represents the state space of the MDP, which contains all possible values of the states. $A$ represents the action space of the agent and $P_{XA}$ represents the state transition probability for state $X$ and action $A$. $P_{XA}$ gives the distribution of the next state if the state of the environment is $X$ and the action of agent is $A$. $\gamma$ represents the discount rate, which takes values between 0 and 1. $R$

represents the reward that the agent receives after taking an action.

As shown in Fig. 10, the RL allows the agent to interact with the environment to generate trajectories. In the energy management problem, a complete trajectory starts at the first step of the driving cycle and ends at the last step. These trajectories consist of a sequence of decisions. Except for the end, each decision introduces the following information: the state $\xi_t$ of the previous step of the environment, the action $u_t$ made by agent based on state, the reward $r_t$ obtained by the action and the next state $\xi_{t+1}$. For the current state $\xi_t$, its state value $V_\pi(\xi_t)$ can be expressed as:

$$V_\pi(\xi) = E_\pi\left[\sum_t^\infty \gamma^t r_t \,|\, \xi_t = \xi\right] \quad (18)$$

where the value of $\gamma$ is close to 0 will make the reward expectation focuses on the short-term gain and the value of $\gamma$ is close to 1, the reward expectation focuses on the long-term gain. The objective function $J_\pi$ of the reinforcement learning algorithm, is to iteratively update the strategy $\pi$ based on the trajectory data so that the reward expectation $V_\pi$ of the current strategy is maximized:

$$J_\pi = \arg\max_\pi V_\pi \quad (19)$$

### 3.2 Algorithms for RL

Model-free reinforcement learning methods are the most important research content in reinforcement learning, which learn the state-action value function through interaction with the environment. Research and improvements around model-free reinforcement learning have profoundly influenced the development of deep reinforcement learning. This paper implements four model-free reinforcement learning algorithms along the developmental path of reinforcement learning, covering three reinforcement learning (TD-learning, On/off policy, N-step) improvement directions.

#### A. Monte Carlo Method（MC）

MC is a reinforcement learning algorithm that provides unbiased state value estimations. The estimation of each state depends on the reward received by the agent during the time period $k$ from the current time step $t$ to the end of the iteration. The cumulative return $G_t$ can be expressed as:

$$G_t = r_{t+1} + \gamma r_{t+2} + \gamma^2 r_{t+3} + \cdots + \gamma^k r_{t+k+1} \quad (20)$$

In MC, estimation $V_\pi(\xi_t)$ of the state $\xi$ by the current strategy can be expressed as the mean value of the cumulative rewards $G_t$:

$$V_\pi(\xi) = E_\pi\left[G_t \,|\, \xi_t = \xi\right] \approx \frac{\sum_{i:\xi_t^i = \xi}^N G_t^i}{N} \quad (21)$$

where $\sum_{i:\xi_t^i = \xi}^N G_t^i$ denotes the sum of the cumulative rewards $G_t$ for each visit to state $\xi_t$, $N$ is the number of visits to the current state $\xi_t$. According to the law of large numbers, the exact expectation value of the function can be obtained when $N$ approaches infinity. The update process of its state value is as follows:

$$V_\pi(\xi_t) = V_\pi(\xi_t) + \alpha\left(G_t - V(\xi_t)\right) \quad (22)$$

Combining (20), (21), (22) for analysis, the MC approach does not fully utilize the structure of the MDP to update the state value based on the trajectory at each moment. Instead, it uses the whole exploration trajectory for the state value function to update. While this approach to strategy updating provides accurate and unbiased state value estimations, it reduces the frequency of strategy updating.

#### B. Q-learning and SARSA

Q-learning and SARSA as temporal difference methods fully use the MDP structure to update state value in a recursive form for each node in the decision sequence. They balance the exploration and exploitation of the trajectory data and achieve a faster speed of strategy updating. The state value update form of the temporal difference method is shown as follows:

$$V_\pi(\xi_t) = V_\pi(\xi_t) + \alpha\left(r_{t+1} + \gamma V_\pi(\xi_{t+1}) - V_\pi(\xi_t)\right) \quad (23)$$

where $\alpha$ is the update step-size of the policy evaluation, often referred to as the learning rate; $r_{t+1} + \gamma V_\pi(\xi_{t+1})$ is the correction to the state evaluation of the policy, and $TD_{target}$. $r_{t+1} + \gamma V_\pi(\xi_{t+1}) - V_\pi(\xi_t)$ is the update deviation of the evaluation, referred to as $TD_{error}$.

SARSA, as an On-policy algorithm, has a policy update process consistent with the behavioral policy, and its learning target is called $TD_{target}$:

$$TD_{target} = r_t + \gamma Q_\pi(\xi_{t+1}, u_{t+1}) \quad (24)$$

where $Q_\pi$ denotes the action-state value function of the current behavioural strategy $\pi$, which can be expressed as:

$$Q_\pi(\xi, u) = E_\pi\left[\sum_t^\infty \gamma^t r_t \,|\, \xi_t = \xi, u_t = u\right] \quad (25)$$

For the trajectory $(\xi_t, u_t, r_t, \xi_{t+1}, u_{t+1})$, SARSA has the following update process for action-state value function $Q_\pi$.

$$Q_\pi(\xi_t, u_t) = Q_\pi(\xi_t, u_t) + \\ \alpha\left(r_t + \gamma Q_\pi(\xi_{t+1}, u_{t+1}) - Q_\pi(\xi_t, u_t)\right) \quad (26)$$

The On-policy algorithms represented by SARSA use state-action value of the next-step to update when updating $Q_\pi$. Q-Learning uses optimal value estimation instead of behavioral strategy estimation to calculate $TD_{target}$.

$$TD_{target} = r_t + \gamma \max_u Q(\xi_{t+1}, u) \quad (27)$$

$$Q_\pi(\xi_t, u_t) = Q_\pi(\xi_{t+1}, u_t) + \\ \alpha\left(r_t + \gamma \max_u Q(\xi_{t+1}, u_t) - Q_\pi(\xi_t, u_t)\right) \quad (28)$$

where $Q(\xi_{t+1}, u_t)$ denotes the highest state estimation of $\xi_{t+1}$. Because the control signal corresponding to max state value is often not exactly the same as the trajectory information of the behavioural policy, Q-learning is considered as a classical Off-policy approach.

*C. SARSA-Lambada*

In addition to the policy iteration methods with single-step update, there is also policy iteration methods based on the previous n-steps trajectories represented by SARSA-Lambda to reduce the evaluation bias of the state in TD-learning methods, and the state evaluation update function is:

$$V_\pi(\xi_t) = V_\pi(\xi_t) + \alpha \left( G_t^{(n)} - V(\xi_t) \right) \tag{29}$$

The cumulative reward of n-steps is $G_t^{(n)}$, which takes the form as:

$$G_t^{(n)} = r_{t+1} + \gamma r_{t+2} + \cdots + \gamma^{n-1} r_{t+n} + \gamma^n V(\xi_{t+n}) \tag{30}$$

where $n$ denotes the step-size of the reward accumulation. The exact value of $n$ is challenging to specify, and TD-learning converges to the MC method as $n$ approaches infinity. $TD(\lambda)$ introduces a trace decay factor $\lambda$ and a qualification trajectory $E_t(\xi)$ [41]:

$$E_t(\xi) = \begin{cases} \gamma\lambda E_{t-1}, \text{if } \xi \neq \xi_t \\ \gamma\lambda E_{t-1} + 1, \text{if } \xi = \xi_t \end{cases} \tag{31}$$

The state value function of $TD(\lambda)$ is then expanded to the following form:

$$V_\pi(\xi_t) = V_\pi(\xi_t) + \alpha \left( r_{t+1} + \gamma V_\pi(\xi_{t+1}) - V_\pi(\xi_t) \right) E_t(\xi) \tag{32}$$

This paper presents the application of $TD(\lambda)$ in RL-EMS using SARSA-Lambda as an example, and the update process is as follows.

$$Q_\pi(s_t, a_t) = Q_\pi(s_t, a_t) + \\ \alpha \left( r + \gamma Q_\pi(s_{t+1}, a_{t+1}) - Q_\pi(s_t, a_t) \right) E_t(s) \tag{33}$$

The impact of these four algorithms on RL-based EMS is analyzed in detail in the section 4.1 of this paper.

*3.3 State and Action in RL Algorithms*

As an AI-driven MPS-EV energy management strategy, the state space and action space of reinforcement learning describe the perception and decision-making capabilities of the energy management. In past EMS studies, speed, acceleration, power demand of the vehicle and battery SOC have been frequently selected to describe the vehicle state. Considering that the vehicle power demand is a comprehensive representation of the current load, speed, and acceleration, this paper adopts the vehicle power demand and SOC as vehicle states to represent the dynamics and powertrain characteristics of the MPS-EV during driving. For PHEV, the torque distribution ratio between engine and motor is chosen as the action space, which takes in the range [0, 1]. For FCEV, the output power of the fuel cell is chosen as the action space, which takes values in the range [0, 55] in kW.

This paper does not focus on further elaboration of the design of states and actions. The different demands of the vehicle models and control tasks lead to a large gap in action and state space design between related studies, the differences in performances due to this gap are not related to the principle of reinforcement learning. However, the discrete granularity of states and actions will cause dramatic changes in the state transition of the MDP. This paper investigates the impact of state and action space discretization on RL-EMS solutions with discrete granularity taking values in the range [5, 11, 21, 31, 41, 51, 61, 71, 81, 91, 101]. The differences in energy consumption due to perception and decision granularity is discussed in Section 4.2.

*3.4 Reward Function in RL Algorithms*

The reward function of reinforcement learning directly reflects the optimization objective of MPS-EV energy management task. Past studies have commonly used the minimizations of fuel consumption and equivalent fuel consumption as objectives to set the reward function.

*A. Minimization of fuel consumption*

The minimization of fuel consumption is an intuitive reward function setting that calculates the current reward of RL-EMS. The fuel consumption minimization reward function $r_{PHEV}$ of PHEV and the hydrogen fuel consumption minimization reward function $r_{FCEV}$ of FCEV can be expressed as:

$$r_{PHEV} = \tau - \dot{m}_{fuel} \tag{34}$$

$$r_{FCEV} = \tau - \dot{m}_{H_2} \tag{35}$$

where $\dot{m}_{fuel}$ is the instantaneous fuel consumption mass in grams, $\dot{m}_{H_2}$ is the instantaneous hydrogen consumption mass in grams, and $\tau$ is a constant, which is 1 in this paper. This reward function is used in the section 4.4 as an important indicator to evaluate the performance of energy consumption optimization under different experimental variables.

*B. Minimization of equivalent fuel consumption*

Minimizing the equivalent fuel consumption has become an important optimization objective in EMS research, which considers the fuel consumption during driving and adds the electricity consumption to the objective function through an equivalence factor. This paper summarizes the current MPS-EV equivalent fuel consumption reward: 1) the equivalent fuel consumption reward based on the instantaneous electricity consumption $r_{eqi}$. 2) the equivalent fuel consumption reward based on the current overall electricity consumption $r_{eqt}$.

$$r_{eqi} = \tau - \left( \dot{m}_F + \alpha\Delta\xi_t \right) \tag{36}$$

$$r_{eqt} = \tau - \left( \dot{m}_F + \beta \left( \Delta\overline{\xi}_t \right)^2 \right) \tag{37}$$

where $\dot{m}_F$ denotes the current instantaneous fuel/hydrogen consumption and $\Delta\xi_t$ is the instantaneous electricity consumption, describing the SOC variation between two time steps.

$$\Delta\xi_t = \xi_t - \xi_{t+1} \tag{38}$$

The current overall electricity consumption describes the amount of SOC change compared with the initial state.

$$\Delta\overline{\xi}_t = \xi_0 - \xi_t \tag{39}$$

The common forms of energy consumption parameters $\alpha$ and $\beta$ are:

$$\alpha = \beta = S \frac{3600 V_{Bat} Q_{max}}{Q_{LHV}} \tag{40}$$

where $V_{Bat}$ denotes the open-circuit voltage of the battery, $Q_{max}$ denotes the battery capacity, $Q_{LHV}$ denotes the fuel heat value, and $S$ denotes the equivalence factor. In this paper, the differences in energy consumption optimization between two types of reward functions with different equivalence factors are discussed in detail in section 4.4, which demonstrates the effect of the variation of the reinforcement learning optimization objective on the EMS charging and discharging process.

### 3.5 Hyper-parameters

Reinforcement learning is sensitive to hyperparameters, and the main hyperparameters of RL-EMS include learning rate, exploration rate, discount rate, training period, and SOC starting point. The learning rate represents the step-size of updating for the action-state value of the current policy. The significant learning rate will get drastic updates of each state value are and the smaller the value will lead the trajectory updates to conservative. To balance exploration and exploitation, the behavioral strategy uses the $\varepsilon$-greedy method, with $\varepsilon$ representing the exploration rate. This allows each state to randomly select control signals from the action space with a probability of $\varepsilon$. A larger exploration rate means more explorations at random. The discount rate takes values in the range [0, 1] and is used to calculate the cumulative rewards. The small discount rate will make the recent returns valued in the state value update. The number of training episodes describes the length of the reinforcement learning training cycle, and the larger the value, the more trajectories can be provided for updating. The SOC starting point represents the initial SOC at each process for trajectory collection, and the value reflects the impact of SOC on trajectory collection and policy updating. In section 4.3, this paper discusses the effects of the learning rate, exploration rate, discount rate, training number, and SOC starting point on RL-EMS. Table 3 shows the parameter selection range of hyperparameters.

TABLE III
Hyper-parameters.

| Hyper-parameters | Value |
|---|---|
| Learning Rate | [0.0005, 0.03, 0.3] |
| Explore Rate | [0.3, 0.6, 0.9] |
| Discount | [0.495, 0.995] |
| Train Episodes | [2000, 5000, 10000] |
| Start SOC | [0.49, 0.59, 0.65] |

### IV. EXPERIMENT AND RESULTS

All models and algorithms in this paper are implemented through Python and executed in a computing workstation with two Intel E5 26330V4 and 256GB RAM. Based on the MPS-EV model, this paper introduces the Worldwide harmonized Light duty driving Test Cycle (WLTC) [42] to simulate the working condition of driving. This section introduces different algorithms, discretization granularity, hyperparameters, and reward functions to develop different RL-EMS solutions. The key factors of RL-EMS performance are analyzed and pointed out by calculating metrics such as average energy cost, powertrain efficiency, and SOC trajectory during the driving cycle.

### 4.1 Differences between Algorithms

This paper implements four classical reinforcement learning algorithms: the Monte Carlo method, Q-learning, SARSA, and SARSA-Lambada. Each algorithm trains the RL-EMS behavioural strategy of PHEV and FCEV by 20,000 iterations in WLTC, and the strategies are evaluated every hundred times with an exploration rate of 0.3. The energy cost, SOC trajectories, and energy utilization efficiency are recorded, and these interim results describe the searching process for optimization of the four reinforcement learning algorithms.

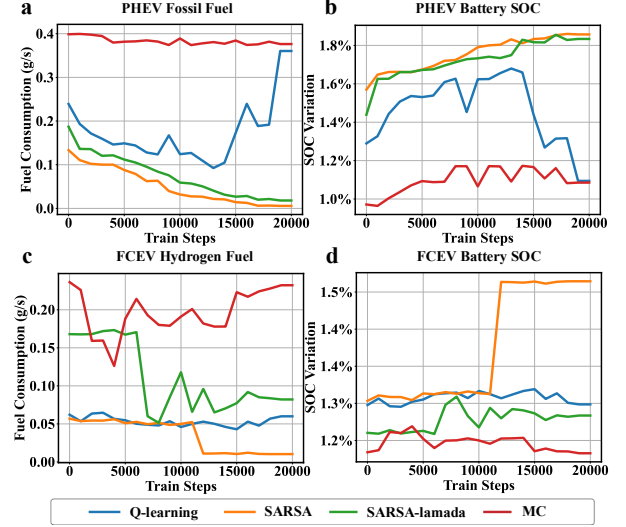

Fig. 11. The average energy costs of different RL algorithms on RL-EMS training.

As shown in Fig. 11, the MC-based EMS has the highest average energy cost per time step of all solutions during the training process of the EMS based on the PHEV model by the RL algorithm. SARSA-based EMS shows a very extreme discharge strategy with a low average energy cost per time step but a high-power consumption of more than 1.8%. These results show that RL-based EMSs have learned more about how to use the electricity saved by the battery to meet the vehicle power demand. SARSA-Lambada shows a similar training curve to SARSA, but its discharge strategy is relatively moderate.

Fig. 12 shows SOC variations of RL-EMS over 2000 step iterations, the magenta curves represent the SOC trajectory of the RL-EMS at the beginning of the training process, while the green curves represent the SOC trajectory of the RL-EMS at the end of the training process, and the different algorithms present various styles of optimization processes. MC generates a different SOC trajectory from the TD method in the first half of the driving cycle. The subsequent SARSA-based EMS and SARSA-Lambada-based EMS exhibit extreme discharge processes. Their On-policy algorithm-based strategies release 20% of the battery power and violate the SOC state constraint within 1000-1500 simulation time steps. It is evident that SARSA algorithm is trapped into the local optimum for single-step fuel consumption, while Q-learning, as an Off-policy TD method, learns to reduce fuel consumption by using discharge and jumps out of the local optimum to achieve the optimal solution under the state constraint according to the penalty for violation the SOC safety boundary.

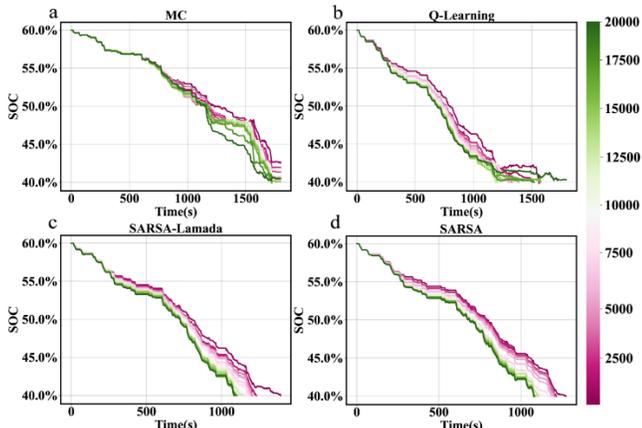

**Fig. 12.** The variations trajectories of SOC during RL training in PHEV.

The engine working points further demonstrate the difference between MC and temporal difference algorithms. The improvement of n-steps to temporal difference methods is also presented. As shown in Fig. 13, the final results of both MC and Q-learning methods generate many engine operating points. However, the MC-based RL-EMS is not efficient for engine utilization with engine operating points concentrated in the red-deep high fuel consumption rate zone, proved by the two-dimensional projection of engine fuel consumption efficiency on engine speed and torque in Fig.13. The SARSA-based RL-EMS concentrates the engine utilization in the low torque range, which leads to extremely high fuel consumption rate and is an extreme engine torque distribution strategy. The RL-EMS based on SARSA-Lambada, compared to the SARSA method, concentrates the engine operating points in the lower fuel consumption rate zone and better balances the torque distribution between the engine and the motor.

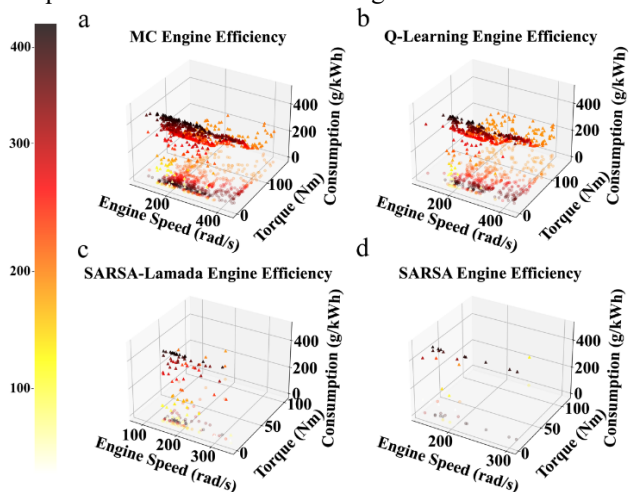

**Fig. 13.** ICE efficiency in the simulation at the end of 20,000 iterations: (a) Engine Efficiency in MC-based EMS; (b) Engine Efficiency in Q-learning-based EMS; (c) Engine Efficiency in SARSA-Lambada-based EMS; (d) Engine Efficiency in SARSA-based EMS.

These four types of algorithms are also applied to the RL-EMS study of FCEV, where the optimization objective is changed from fuel consumption minimization to hydrogen consumption minimization, and the control signal is the fuel cell output power. As shown in Fig. 11, MC-based EMS remains the solution with the highest single-step hydrogen consumption. SARSA remains the solution with the lowest single-step hydrogen consumption and the most extreme discharge strategy during 20,000 training cycles. Fig. 15 shows the variation of the SOC trajectory of RL-EMS during the FCEV training process, which further confirms that even if the models are different, MC methods still maintain different SOC trajectories from the temporal difference methods. SARSA is still trapped in a local optimum, resulting in incomplete working conditions due to the state limits. Unlike the results based on PHEV, the increase of SARSA -Lambada relative to SARSA is more obvious in FCEV.

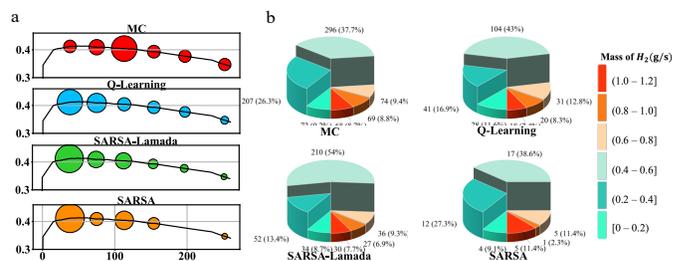

**Fig. 14.** Fuel cell operating efficiency and hydrogen consumption in the simulation at the end of 20,000 policies update iterations.

Fig. 14 (a) illustrates the operating point-efficiency distribution and hydrogen consumption distribution of RL-EMS in the fuel cell, and Fig. 14 (b) depicts the segmented statistics of hydrogen consumption mass in the WLTC working condition. Q-learning makes the best use of the fuel cell compared with other methods. It outputs the power of the fuel cell more efficiently with fewer times of consuming hydrogen.

As shown in Eqs. 24, 27, Q-learning constructs $TD_{target}$ based on the optimal estimation of the next state under policy $Q_\pi$ while SARSA constructs $TD_{target}$ based on the estimation of the next state under policy $Q_\pi$. However, when the current behavioural strategy generates trajectory data for the energy management problem, the current behavioural strategy is not always able to choose the optimal action due to the influence of stochastic exploration and policy performance. Therefore, SARSA, which relies heavily on the quality of the trajectories generated by the current policy, is prone to fall into local optimum. The Q-learning algorithm updates $Q_\pi$ with the optimal estimation of $Q_\pi$, which makes it easier to seek the optimal solution of the policy in Q-learning. As a temporal difference method, Q-learning utilizes the trajectory data at each moment to update the strategy, which greatly improves the strategy development efficiency of RL-EMS.

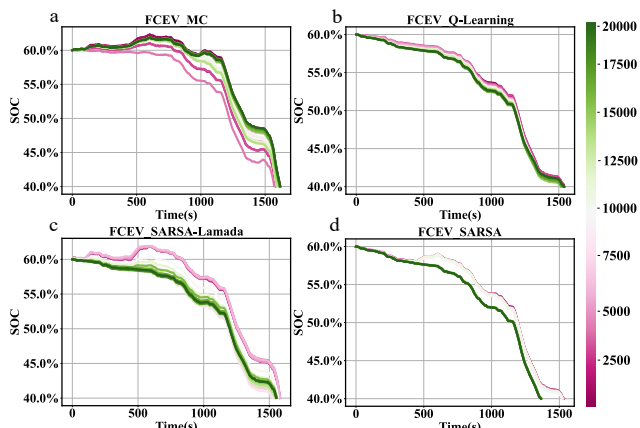

**Fig. 15.** The variations of SOC trajectory during the RL training in FCEV.

### 4.2 Perception and Decision Granularity

It is a common solution for the control problems to discretize the continuous perception and decision space into a collection of discrete values of states and actions. The discrete granularity of the perception and decision space would affect the state transition in control systems, as insufficient girds would lead to unreasonable state transition, while too many girds would lead to high search costs. The impact of discrete granularity on general control problems is not the focus of this paper. However, the impact of discrete granularity of perception and decision space on RL in EMS problems will be discussed in this section.

Fig. 16 shows the SOC trajectories of the RL-EMS for PHEV with perception grids of [11, 21, 51] and decision grids of [5, 11, 21]. During the WLTC working condition, the solutions with a high granularity of perception and decision achieve smoother SOC trajectories, while the solutions with a low granularity of perception and decision show greater discharge amplitudes. The SOC trajectories have the same trend during the peak energy consumption interval, but their discharge amplitudes differ significantly. These results indicate a similar decision propensity of the RL-EMS algorithms during this interval, but the gap in decision granularity causes significant differences among SOC states. This paper selects 11 different grid numbers of perception and decision to further investigate the difference between EMS with fewer or more grids in energy cost. Section 4.1 shows the superiority of the Q-learning algorithm in energy management problems. This section uses the Q-learning algorithm to construct the RL-EMS with 10,000 rounds of iterative training for the PHEV and FCEV models. Its single-step energy consumption results at the end of the iteration are shown in Fig. 18. State and action spaces with low discrete granularity have significant advantages in reducing fuel consumption but consume more power. State and action spaces with high discrete granularity have low single-step power consumption but consume more fuel. Strategies with low discrete granularity are more efficient in engine utilization, and strategies with high discrete granularity are more likely to generate operating points with low efficiency.

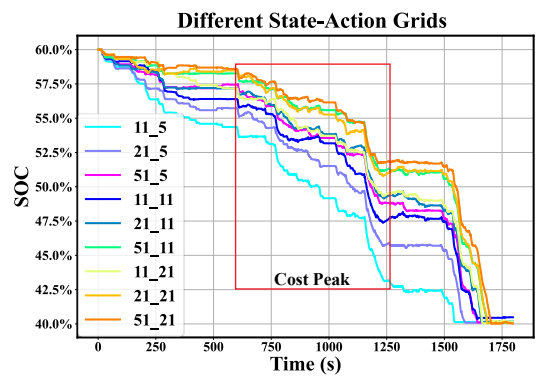

**Fig. 16.** The SOC variations in different perception and decision grids for PHEV. "11_5" represents the SOC trajectory of RL-EMS in simulation after 10,000 iterations. Its state grid is 11 and the action grid is 5.

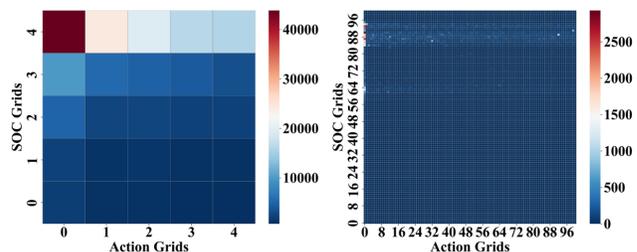

**Fig. 17.** The update counts of Q-table after 100 policy iterations in FCEV.

The reason for this result comes from the fact that state transition is more complex with the setting of high discrete granularity. The basic principle of reinforcement learning is to update the Q-table, which provides the state-action value of the policy by reward r observed from the trajectory. From (21), (23), and (29), if the more frequently the Q value is updated, the state-value estimation by Q-table will be accurate. This makes the average update counts of each state-action value of high discrete granularity lower than that of the low discrete granularity after the same number of iterations, which leads to a high deviation of the state estimation. Fig. 17 demonstrates that the update counts per state-action value are much higher for the small state-action grid number setting than for the lager state-action grid number setting.

The choice of grid number should also consider the physical significance of the optimization objective of the EMS. In the energy management problem regarding equivalent fuel consumption as the optimization objective, the power source distribution should consider fuel consumption and electricity consumption. In the perception and decision space, a large state-action grid number can capture the power variation more accurately and make precise power distribution commands than a small state-action grid number. Fig. 19 shows the equivalent energy consumption levels for various discrete granularity settings of the perception and decision space for a given equivalent fuel consumption target. It demonstrates that the high discrete granularity setting performs much better than the low discrete granularity setting in the equivalent energy optimization problem.

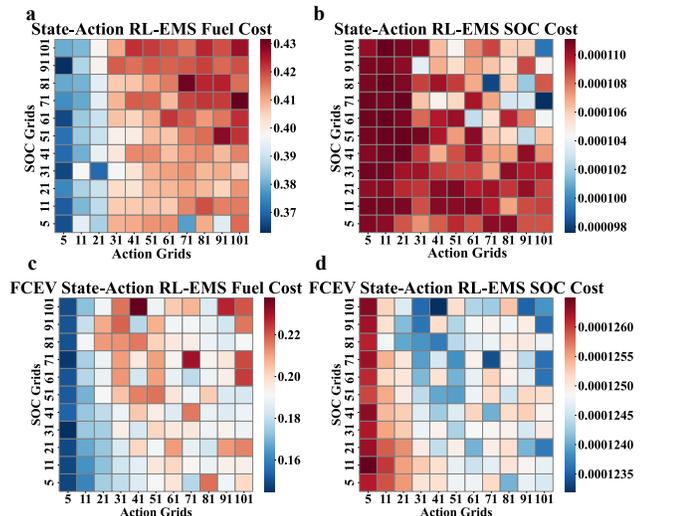

**Fig. 18.** The average energy cost heatmaps in different state-action grid settings.

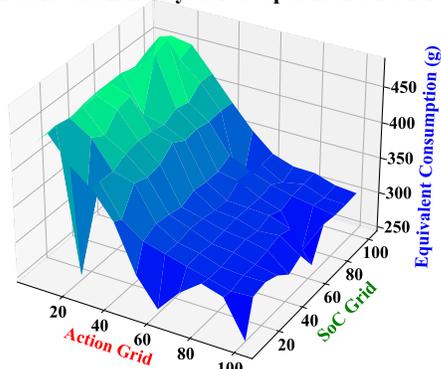

**Fig. 19.** The equivalent energy cost map in different state-action grid settings in PHEV.

### 4.3 Hyper Parameters

When reinforcement learning is tried for application in energy management problems of MPS-EV, hyperparameters come from three main aspects: 1) reinforcement learning parameters, 2) deep reinforcement learning parameters, and 3) parameters of MPS-EV-related settings. The settings and hyperparameters of deep reinforcement learning are strongly related to the deep reinforcement learning algorithm and different neural network settings can affect the training effect. Therefore, the analysis of hyperparameters of deep reinforcement learning is not the content of this paper. The settings related to the reinforcement learning parameters and MPS-EV are summarized in Table 3. This paper applies WLTC as the driving conditions for RL-EMS training in PHEV and FCEV based on the Q-learning algorithm and gives performance analysis.

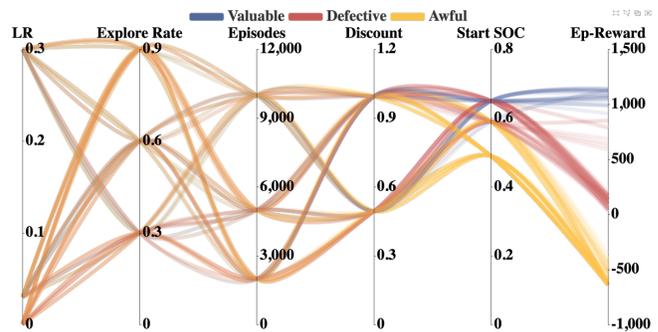

**Fig. 20.** Final sum of rewards at different hyper-parameters setting for RL-EMS in PHEV.

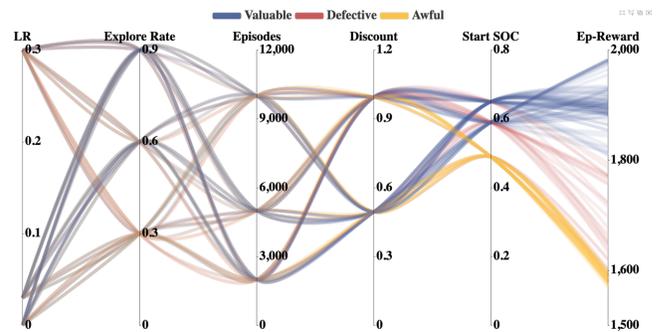

**Fig. 21.** Final sum of rewards at different hyper-parameters setting for RL-EMS in FCEV.

Fig. 20, 21 show the performance of RL-EMS under different hyperparameter settings and vehicle models at the end of the training iterations, respectively. The training results are classified according to the data density of the cumulative rewards: valuable, defective and awful. The valuable results show that this RL-EMS is adequately trained in the experiment and achieved good cumulative rewards results. In contrast, the awful experimental results imply that the hyperparameters of the experiment are not set reasonably, and the experimental results fail to achieve any energy optimization. The defective experimental results represent unsatisfactory energy optimization strategies, which may require more experiments to update the parameters. Among them, the availability of valuable experimental results is marked as 1, the availability of poor experimental results is marked as -1, and the availability of defective experimental results is marked as 0. It can be seen from the parallel coordinate system that the trajectories of different hyperparameters are most obviously transformed on the start SOC coordinate axis. The effects of different hyper-parameters on training can be sorted by calculating the distribution of the experimental trajectories of the two vehicle models on each coordinate axis. Fig. 22 shows the ranking results of the effects of various hyperparameters on RL-EMS performance in the PHEV and FCEV energy management problems. The availability index is close to 1.0, indicating that the value of this parameter is helpful for RL-EMS training. On the contrary, if the availability index is close to - 1.0, it indicates that the value of this parameter is not conducive to training. It can be seen from the experimental results that a low SOC starting point almost means failure of training, while high SOC starting point, appropriate exploration rate, lower

learning rate, and high iteration number are easier to train available EMS strategies. Whether in PHEV or FCEV, the high starting point of SOC can bring a better RL-EMS training effect. The high starting point of SOC for RL-EMS helps the behavioral strategies generate trajectories adequately and safely without frequent penalties for exceeding the constraint due to too low SOC starting point settings. The high number of iterations ensures the amount of training data and the number of policy updates. The low learning and exploration rates ensure the stable update of strategies during the exploration and exploitation of reinforcement learning.

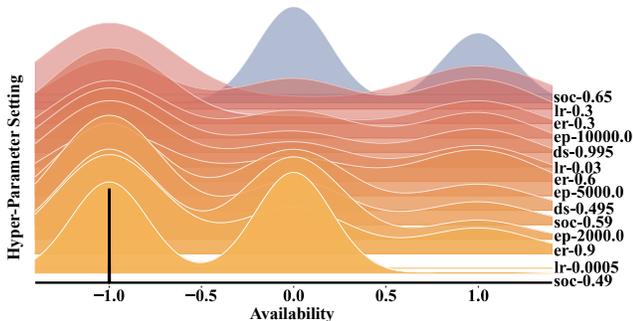

**Fig. 22.** Ranking of influence of hyper-parameters on RL-EMS solution.

### 4.4 Optimization Targets

The reward function guides the optimization direction of RL in the energy management problem, where common reward function settings are fuel consumption minimization and equivalent fuel consumption minimization. The reward function based on fuel consumption minimization is obtained by directly calculating the fuel consumption as in (34) and (35). In Section 3.4, this paper summarizes two currently widely adopted reward functions: the equivalent energy reward $r_{eqi}$ which considers instantaneous energy consumption, and the equivalent energy reward $r_{eqt}$, which considers current overall energy consumption. They calculated the equivalent fuel consumption from microscopic and macroscopic power losses, respectively. The equivalence factor $S$ influenced the calculation process as shown in Eqs. 36, 37. This section discusses the parameter design of the reward function for RL in the energy management problem of MPS-EV by setting different equivalence factors and equivalent fuel consumption functions. The effect of the reward function parameter design is specified by analyzing the discharge process of RL-EMS in PHEV and FCEV under different settings. The reward function parameters and experimental results are shown in Table 4: experiments numbered 0-4 are based on the reward function $r_{eqi}$, and they calculate the equivalent fuel consumption based on the amount of SOC variation at each time step with different equivalence factor $S_\alpha$; experiments numbered 5-9 are based on the reward function $r_{eqt}$, and they calculate the equivalent fuel consumption based on the total SOC variation with different equivalence factor $S_\beta$.

TABLE IV
Different reward setting results.

| Trial No. | | 0 | 1 | 2 | 3 | 4 |
|---|---|---|---|---|---|---|
| $S_\alpha$ | | 1 | 1.5 | 2 | 2.5 | 3 |
| PE HV | Fuel Consum ption | 565.02 | 568.55 | 576.63 | 844.93 | 927.99 |
| | SOC Variatio n | 0.1961 | 0.1974 | 0.1984 | 0.0519 | -0.0043 |
| FC EV | H-Fuel Consum ption | 513.75 | 517.87 | 521.19 | 523.43 | 528.57 |
| | SOC Variatio n | 0.3545 | 0.3539 | 0.3531 | 0.3525 | 0.3521 |
| Trial No. | | 5 | 6 | 7 | 8 | 9 |
| $S_\beta$ | | 0 | 250 | 500 | 750 | 1000 |
| PE HV | Fuel Consum ption | 561.52 | 594.23 | 786.68 | 846.75 | 864.85 |
| | SOC Variatio n | 0.1977 | 0.1966 | 0.0988 | 0.0520 | 0.0400 |
| FC EV | Hydrogen Fuel Consum ption | 510.23 | 655.30 | 661.01 | 688.40 | 717.49 |
| | SOC Variatio n | 0.3565 | 0.3165 | 0.3023 | 0.2736 | 0.2726 |

#### A. PHEV

It can be seen from Fig. 23 (a): that the value of the equivalence factor $S_\alpha$ increases gradually from experiment 0 to experiment 3, and the SOC trajectory changes dramatically during the simulation as the value of $S_\alpha$ increases. When the equivalence factor $S_\alpha$ takes the value below 2, the SOC trajectory keeps a similar trend. When the equivalence factor $S_\alpha$ takes a value of 2.5, RL-EMS reduces the total power consumption during the WLTC working condition. There is no longer an intensive discharge process in the SOC trajectory, and an obvious charging process starts to appear; the SOC trajectory shows an obvious SOC maintenance process during the driving cycle when the equivalence factor $S_\alpha$ takes a value of 3, which indicates that this RL-EMS achieves a good power maintenance strategy for the WLTC working condition. Considering the engine working points, as shown in Figure 24, the training result of number 4 uses the engine more frequently. These results introduce that the effect of the equivalence factor $S_\alpha$ on the RL-EMS is nonlinear and very sensitive when the reward function $r_{eqi}$ is used for the RL-EMS. Since such a reward function considers only the instantaneous SOC variation rather than the current SOC state. The reward is calculated by specifying the weight of fuel and electricity, resulting in differences between the RL-EMS strategies. This tends to result in a problem that if $S_\alpha$ is too high or too low, the SOC will keep rising or falling until the constraint is exceeded, at which point the reward function needs to incorporate an additional penalty function to ensure the correct estimation of the dangerous behaviour strategy by reinforcement learning.

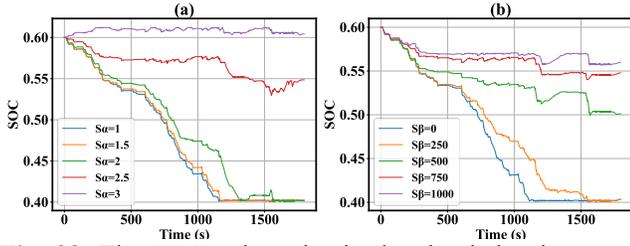

**Fig. 23.** The SOC trajectories in the simulation is set under different reward functions after 10,000 iterations.

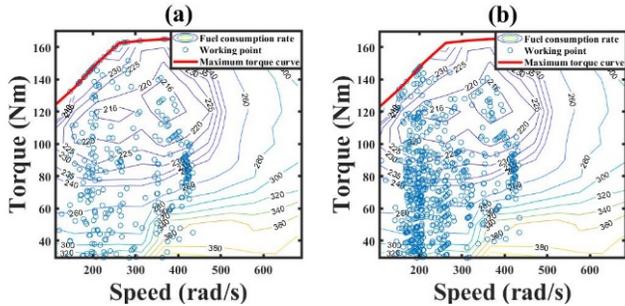

**Fig. 24.** ICE operation points in different $S_\alpha$ settings: (a) $S_\alpha$: 2;(b) $S_\alpha$: 3.

Fig. 23(b) shows the SOC trajectories generated in the simulation after 10,000 iterations, with $S_\beta$ taking values that gradually increase from 0 to 1000. It is worth noting that these SOC trajectories hardly differ from 0 to 200 seconds because at the beginning of the driving cycle, $\Delta\bar{\xi}_t$ is too small, and the modulation effect of $S_\beta$ is low. In the subsequent process, RL-EMS always tries to stabilize the SOC around a certain value, making these SOC trajectories have similar trends. The results show that increasing $S_\beta$ makes RL-EMS less likely to produce dangerous strategies that violate the state constraints. The RL-EMS based on the reward function $r_{eqt}$ is not sensitive to the parameter $S_\beta$.

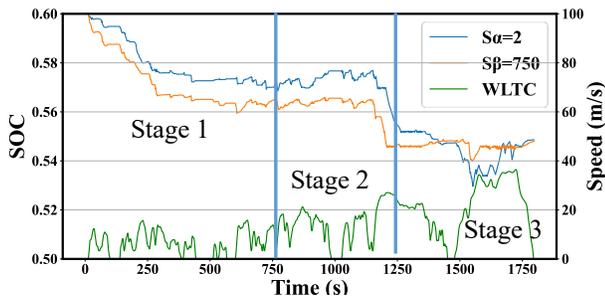

**Fig. 25.** Compare two reward function settings with similar energy consumption in PHEV.

The differences between the two types of reward functions are analyzed by taking "$S_\alpha = 2$" and "$S_\beta = 750$"as examples. As shown in Fig. 25, the final SOC of both examples are similar, and the SOC trends show different characteristics in the three stages of the whole working condition.

In State 1, the SOC variation of "$S_\beta = 750$" decreases rapidly at the first peak of power consumption because the SOC changes are small at this time, and the reward function $r_{eqi}$ does not play a significant role in fuel-electricity balance. In contrast, the SOC of "$S_\alpha = 2$" decreases more slowly, avoiding electricity cost. In State 2, the SOC variations are similar for both, representing that the value of $\beta\left(\Delta\xi(t)\right)^2$ in the reward function $r_{eqi}$ increases as the SOC decreases. At this time, the reward function achieves a reasonable weight distribution between electricity and fuel. In the later stage of Stage 2, the SOC variation of "$S_\beta = 750$" tends to remain stable or even rise. With the decline of SOC, the reward function $r_{eqi}$-based EMS pays more attention to the maintenance of SOC. In State 3, the rise and fall of SOC of "$S_\beta = 750$" are suppressed, and SOC only decreases to a small extent even in the period of peak power consumption from 1500 to 1700 seconds.

This result shows the fact that the reward function $r_{eqi}$ has a strong SOC maintenance capability compared to the reward function $r_{eqt}$, but reduces the buffering effect of the battery.

### B. FCEV

As shown in Figure 26, the difference in motor and battery capacity results in different SOC trajectories for the RL-EMS strategies in FCEV. Combined with the results in Table 4, the final SOC variation of RL-EMS for the reward function $r_{eqt}$ substantially exceeds 0.3. Only when $S_\alpha = 3$ the RL-EMS increases the battery power reserve by consuming hydrogen fuel during the first 1000 seconds of the driving cycle, but this result still substantially consumes the electricity saved by the battery at the end of the driving cycle. The final SOC variation of the RL-EMS is around 0.3 when the reward function $r_{eqi}$ is applied. In the first 1000 seconds of the driving cycle, the battery power is increased by consuming hydrogen fuel to obtain additional reserves and maintain a higher SOC at the end of the driving cycle.

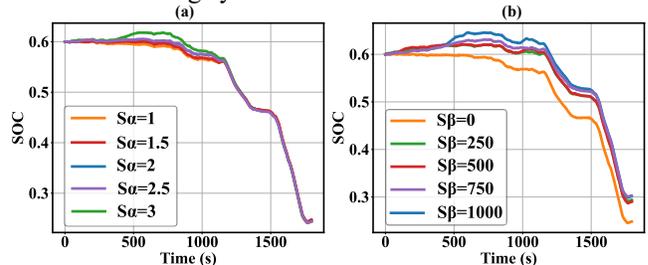

**Fig. 26.** The FCEV SOC trajectories in the simulation is set under different reward functions after 10,000 iterations.

According to the hydrogen fuel consumption results in Table 4, similar to the results in PHEV, the RL-EMS with an optimization objective based on overall power consumption still achieves a more reasonable SOC maintenance effect in FCEV, while the RL-EMS with an optimization objective based on instantaneous power loss closely follows the working condition demand, and the hydrogen fuel consumption is lower.

## V. DISCUSSION

This paper presents an empirical study based on four key factors: different algorithms, perception-decision granularity, reward functions and hyperparameter settings. From the analysis in Section IV, we are able to draw the following conclusions:

(1) Selection of reinforcement learning algorithm. The TD-learning methods have better efficiency in policy updating, and the Q-learning algorithm with Off-policy architecture can help the RL-EMS algorithm achieve optimal results. The multi-step update is worth trying as an optional improvement scheme when RL-EMS performance is poor.

(2) Perception-decision granularity. The density perceptual-decision granularity brings the problem of low update efficiency of Q-table, but it will have some advantages in multi-objective optimization problems. It is crucial to decide the appropriate perceptual-decision granularity, which should be explorable from sparse to dense during the experiment.

(3) Hyperparameters. The initial SOC is the most important hyperparameter and should be selected at the highest value of the reasonable range for training RL-EMS. The iterative training period of RL-EMS should be increased as much as possible, and a longer training process can usually achieve better training results. In addition, our experimental results show that lower learning and exploration rates can improve the final performance of RL-EMS.

(4) Reward functions. The parameters of the reward function represented by the equivalence factor can drastically affect the final performance and SOC trajectory of RL-EMS, when solving multi-objective optimization problems. The experimental results show that the reward function considering the current-overall energy consumption can better control the battery state and not easily enter the restricted interval of SOC. The instantaneous energy consumption-based reward function can follow the power demand more sensitively, but it tends to cause drastic battery state changes and get the safety boundaries.

There are still some limitations in this paper:

(1) Lack of a discussion on deep reinforcement learning. This paper aims to figure out the critical factors of the performance improvement of RL-EMSs. Due to the black-box characteristic of the deep neural network, deep reinforcement learning and its settings are not introduced in this paper.

(2) Robustness issues. This paper discusses two reward functions in a driving cycle (WLTC) in specific settings. There is no analysis of the robustness issues of RL-EMS. More driving cycles, working conditions and object functions should be introduced to benchmark RL-EMS performance.

For these reasons, more factors and settings, especially deep learning-related parameters, will be thoroughly analyzed and discussed in future research. More data on working conditions and simulated environments close to real-world driving cycles will be used for RL-EMS development and performance analysis. The important factors affecting the adaptability of RL-EMS will be revealed in the actual driving data.

## VI. CONCLUSION

This paper describes the general optimization process of RL in energy management problems and discusses the effects of algorithms, perception and decision granularity, hyperparameters, and reward functions on RL-EMS by performance evaluation on two typical MPS-EV models, PHEV and FCEV as study cases. The proposed analysis method can be theoretically applied to different configurations of MPS-EVs. Even though the RL-EMS will perform differently in different vehicles, the analysis results are general in nature, and the trends observed in the experiment do not change.

The results of this paper can provide a crucial basis and guidance for the development of RL-EMS in MPS-EVs, which can promote further applications of RL in energy management problems. In order to reduce the experimental cost of empirical studies, this paper does not discuss in depth the application of deep reinforcement learning in MPS-EVs and the robustness of RL-EMS. Deep reinforcement learning-based energy management strategies (DRL-EMSs) further improve the energy-saving performance of RL-EMS while also bringing more configuration issues such as network structure design, algorithm architecture design and more hyperparameter selection. The robustness of RL-EMS still needs further validation and analysis to apply the RL-EMS in actual driving. Therefore, the empirical analysis of reinforcement learning in MPS-EV can be further extended to the exploration of deep reinforcement learning and EMS robustness.

## REFERENCE

[1] Pisu, Pierluigi, and Giorgio Rizzoni. "A comparative study of supervisory control strategies for hybrid electric vehicles." IEEE transactions on control systems technology 15.3 (2007): 506-518.

[2] Lachhab, Islem, and Lotfi Krichen. "An improved energy management strategy for FC/UC hybrid electric vehicles propelled by motor-wheels." International journal of hydrogen energy 39.1 (2014): 571-581.

[3] Yun, Haitao, et al. "Energy management for fuel cell hybrid vehicles based on a stiffness coefficient model." International Journal of Hydrogen Energy 40.1 (2015): 633-641.

[4] Farrall, S. D., and R. P. Jones. "Energy management in an automotive electric/heat engine hybrid powertrain using fuzzy decision making." Proceedings of 8th IEEE International Symposium on Intelligent Control. IEEE, 1993.

[5] Lee, Hyeoun-Dong, and Seung-Ki Sul. "Fuzzy-logic-based torque control strategy for parallel-type hybrid electric vehicle." IEEE Transactions on Industrial Electronics 45.4 (1998): 625-632.

[6] Li, Huan, et al. "A review of energy management strategy for fuel cell hybrid electric vehicle." 2017 IEEE Vehicle Power and Propulsion Conference (VPPC). IEEE, 2017.

[7] Chen, Zheng, et al. "Temporal-difference learning-based stochastic energy management for plug-in hybrid electric buses." IEEE

Transactions on Intelligent Transportation Systems 20.6 (2018): 2378-2388.

[8] Enang, Wisdom, and Chris Bannister. "Modelling and control of hybrid electric vehicles (A comprehensive review)." Renewable and Sustainable Energy Reviews 74 (2017): 1210-1239.

[9] Chrenko, Daniela, et al. "Novel classification of control strategies for hybrid electric vehicles." 2015 IEEE vehicle power and propulsion conference (VPPC). IEEE, 2015.

[10] Zhang, Yudong, Shuihua Wang, and Genlin Ji. "A comprehensive survey on particle swarm optimization algorithm and its applications." Mathematical problems in engineering 2015 (2015).

[11] Das, Himadry Shekhar, Chee Wei Tan, and A. H. M. Yatim. "Fuel cell hybrid electric vehicles: A review on power conditioning units and topologies." Renewable and Sustainable Energy Reviews 76 (2017): 268-291.

[12] Zhang, Pei, Fuwu Yan, and Changqing Du. "A comprehensive analysis of energy management strategies for hybrid electric vehicles based on bibliometrics." Renewable and Sustainable Energy Reviews 48 (2015): 88-104.

[13] Xie S, Hu X, Qi S, et al. An artificial neural network-enhanced energy management strategy for plug-in hybrid electric vehicles[J]. Energy, 2018, 163: 837-848.

[14] Zhang R, Tao J, Zhou H. Fuzzy optimal energy management for fuel cell and supercapacitor systems using neural network based driving pattern recognition[J]. IEEE Transactions on Fuzzy Systems, 2018, 27(1): 45-57.

[15] Xi L, Zhang X, Sun C, et al. Intelligent energy management control for extended range electric vehicles based on dynamic programming and neural network[J]. Energies, 2017, 10(11): 1871.

[16] Lin X, Bogdan P, Chang N, et al. Machine learning-based energy management in a hybrid electric vehicle to minimize total operating cost[C]//2015 IEEE/ACM International Conference on Computer-Aided Design (ICCAD). IEEE, 2015: 627-634.

[17] Hou Z, Guo J, Xing J, et al. Machine learning and whale optimization algorithm based design of energy management strategy for plug-in hybrid electric vehicle[J]. IET Intelligent Transport Systems, 2021, 15(8): 1076-1091.

[18] Biswas, Atriya, and Ali Emadi. "Energy management systems for electrified powertrains: State-of-the-art review and future trends." IEEE Transactions on Vehicular Technology 68.7 (2019): 6453-6467.

[19] Sutton, Richard S., and Andrew G. Barto. "Reinforcement learning." Journal of Cognitive Neuroscience 11.1 (1999): 126-134.

[20] Xu, Bin, et al. Real-time reinforcement learning optimized energy management for a 48V mild hybrid electric vehicle. No. 2019-01-1208. 2019.

[21] Lin, Xue, et al. "Reinforcement learning based power management for hybrid electric vehicles." 2014 IEEE/ACM international conference on computer-aided design (ICCAD). IEEE, 2014.

[22] Hu, Yue, et al. "Energy management strategy for a hybrid electric vehicle based on deep reinforcement learning." Applied Sciences 8.2 (2018): 187.

[23] Song, Changhee, et al. "A power management strategy for parallel PHEV using deep Q-Networks." 2018 IEEE Vehicle Power and Propulsion Conference (VPPC). IEEE, 2018.

[24] Liu, Chang, and Yi Lu Murphey. "Power management for plug-in hybrid electric vehicles using reinforcement learning with trip information." 2014 IEEE transportation electrification conference and expo (ITEC). IEEE, 2014.

[25] Yuan, Jingni, Lin Yang, and Qu Chen. "Intelligent energy management strategy based on hierarchical approximate global optimization for plug-in fuel cell hybrid electric vehicles." International Journal of Hydrogen Energy 43.16 (2018): 8063-8078.

[26] Han, Xuefeng, et al. "Energy management based on reinforcement learning with double deep Q-learning for a hybrid electric tracked vehicle." Applied Energy 254 (2019): 113708.

[27] Qi, Xuewei, et al. "Deep reinforcement learning enabled self-learning control for energy efficient driving." Transportation Research Part C: Emerging Technologies 99 (2019): 67-81.

[28] Liu, Teng, et al. "A heuristic planning reinforcement learning-based energy management for power-split plug-in hybrid electric vehicles." IEEE Transactions on Industrial Informatics 15.12 (2019): 6436-6445.

[29] Vagg, Christopher, et al. "Stochastic dynamic programming in the real-world control of hybrid electric vehicles." IEEE Transactions on Control Systems Technology 24.3 (2015): 853-866.

[30] Zeng, Xiangrui, and Junmin Wang. "A parallel hybrid electric vehicle energy management strategy using stochastic model predictive control with road grade preview." IEEE Transactions on Control Systems Technology 23.6 (2015): 2416-2423

[31] van Keulen, Thijs, et al. "Design, implementation, and experimental validation of optimal power split control for hybrid electric trucks." Control Engineering Practice 20.5 (2012): 547-558.

[32] Xu, Fuguo, et al. "Battery‐lifetime‐conscious energy management strategy based on sp‐sdp for commuter plug‐in hybrid electric vehicles." IEEJ Transactions on Electrical and Electronic Engineering 13.3 (2018): 472-479.

[33] Tang, Li, Giorgio Rizzoni, and Simona Onori. "Energy management strategy for HEVs including battery life optimization." IEEE Transactions on Transportation Electrification 1.3 (2015): 211-222.

[34] Stockar, Stephanie, et al. "Energy-optimal control of plug-in hybrid electric vehicles for real-world driving cycles." IEEE Transactions on Vehicular Technology 60.7 (2011): 2949-2962.

[35] Schori, Markus, et al. "Optimal calibration of map-based energy management for plug-in parallel hybrid configurations: a hybrid optimal control approach." IEEE Transactions on Vehicular Technology 64.9 (2014): 3897-3907.

[36] Xiong R, Cao J, Yu Q. Reinforcement learning-based real-time power management for hybrid energy storage system in the plug-in hybrid electric vehicle[J]. Applied energy, 2018, 211: 538-548.

[37] Liu T, Zou Y, Liu D, et al. Reinforcement learning‐based energy management strategy for a hybrid electric tracked vehicle[J]. Energies, 2015, 8(7): 7243-7260.

[38] Kouche-Biyouki, Shahrzad Amrollahi, et al. "Power management strategy of hybrid vehicles using sarsa method." Electrical Engineering (ICEE), Iranian Conference on. IEEE, 2018.

[39] Qi, Xuewei, et al. "Deep reinforcement learning enabled self-learning control for energy efficient driving." Transportation Research Part C: Emerging Technologies 99 (2019): 67-81.

[40] Qi, Xuewei, et al. "A novel blended real-time energy management strategy for plug-in hybrid electric vehicle commute trips." 2015 IEEE 18th international conference on intelligent transportation systems. IEEE, 2015.

[41] Sutton, Richard S., and Andrew G. Barto. Reinforcement learning: An introduction. MIT press, 2018.

[42] Tutuianu, Monica, et al. "Development of a World-wide Worldwide harmonized Light duty driving Test Cycle (WLTC)." Technical Report (2013).

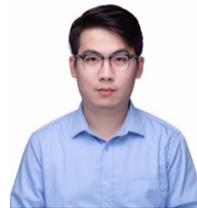

**Jincheng Hu** received the B.E. degree in Information Security from the Tianjin University of Technology, Tianjin, China, in 2019 and the M.Sc. degree in Information Security from the University of Glasgow, Glasgow, UK, in 2022. He is currently working toward the Ph.D. degree in Automotive with the Loughborough University, Loughborough, UK.

His current research interests include reinforcement learning, deep learning, and intelligent system.

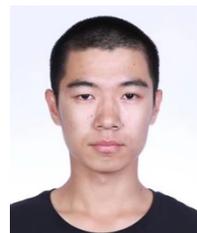

**Yang Lin** received the B.S. degree in Automotive Engineering from Jilin University, Changchun, China, in 2021. He is currently conducting academic research on the Alternative Fuel Vehicle in Jilin University.

His research interests include machine learning, reinforcement learning, energy management strategy for plug-in hybrid vehicles and vehicle steering system.

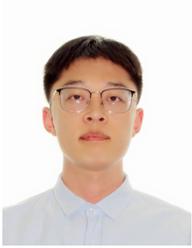

**Jihao Li** received the B.E. degree in Digital media technology from Northeastern University, Shenyang, China, in 2021 and the M.SC. Degree in Artificial Intelligence from the Loughborough University, Loughborough, UK, in 2022. His research interests include deep learning, image recognition and object detection.

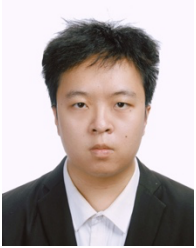

**Zhuoran Hou** received the B.S. degree in Automotive Engineering from Chongqing University, Chongqing, China, in 2017. and the M.S. in Automotive Engineering from Jilin University, China, in 2020. He is currently pursuing continuous academic program involving doctoral studies in automotive engineering with Jilin University, Changchun, China.

His research interests include machine learning, optimal energy management strategy about plug-in hybrid vehicles.

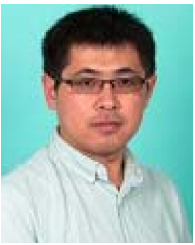

**Dezong Zhao** (Senior Member, IEEE) received the B.Eng. and M.Sc. degrees in Control Science and Engineering from Shandong University in 2003 and 2006, respectively, and Ph.D. degree in Control Science and Engineering from Tsinghua University in 2010. He is currently a Senior Lecturer in Autonomous Systems with the James Watt School of Engineering, University of Glasgow, U.K.

His current research interests include connected and automated vehicles, autonomous vehicles, machine learning, and dynamic optimization.

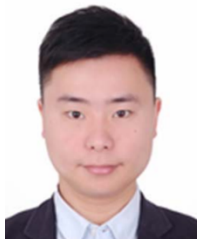

**Quan Zhou** (Member, IEEE) received his BEng and MEng in Automotive Engineering from Wuhan University of Technology, in 2012 and 2015, respectively. He received the Ph.D. in mechanical engineering from the University of Birmingham (UoB) in 2019 that was distinguished by being the school's sole recipient of the UoB Ratcliffe Prize. He is currently Assistant Professor in Automotive Engineering and leads the Research Group of Connected and Autonomous Systems for Electrified Vehicles (CASE-V) at UoB.

His research interests include evolutionary computation, fuzzy logic, reinforcement learning, and their application in vehicular systems.

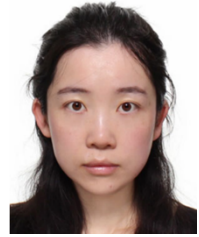

**Jingjing Jiang** (Member, IEEE) received the B.E. degree in Electrical and Electronic Engineering from the University of Birmingham, Birmingham, U.K., and the Harbin Institute of Technology, Harbin, China, in 2010, the M.Sc. degree and the Ph.D. degree from Imperial College London, London, U.K., in 2011 and 2016, respectively, both in Control System.

She has carried out research as part of the Control and Power Group, Imperial College and joined Loughborough University as a Lecturer in September 2018. Her current research interests include driver assistance control and autonomous vehicle control design, control design of systems with constraints, and human-in-the-loop.

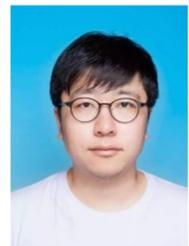

**Yuanjian Zhang** (Member, IEEE) received the M.S. in Automotive Engineering from the Coventry University, UK, in 2013, and the Ph.D. in Automotive Engineering from Jilin University, China, in 2018. In 2018, he joined the University of Surrey, Guildford, UK, as a Research Fellow in advanced vehicle control. From 2019 to 2021, he worked in Sir William Wright Technology Centre, Queen's University Belfast, UK.

He is currently a Lecturer with the Department of Aeronautical and Automotive Engineering, Loughborough University, Loughborough, U.K. He has authored several books and more than 50 peer-reviewed journal papers and conference proceedings.

His current research interests include advanced control on electric vehicle powertrains, vehicle-environment-driver cooperative control, vehicle dynamic control, and intelligent control for driving assist system.